\documentclass[iicol,sn-vancouver,Numbered]{sn-jnl}% Vancouver Reference Style
\usepackage{array}
\usepackage{graphicx}%
\usepackage{multirow}%
\usepackage{float}
\usepackage{amsmath,amssymb,amsfonts}%
\usepackage{amsthm}%
\usepackage{mathrsfs}%
\usepackage[title]{appendix}%
\usepackage{xcolor}%
\usepackage{textcomp}%
\usepackage{manyfoot}%
\usepackage{booktabs}%
\usepackage{algorithm}%
\usepackage{algorithmicx}%
\usepackage{algpseudocode}%
\usepackage{listings}%
\usepackage{rotating}
\usepackage{tabularx}
\usepackage{makecell}
\usepackage{threeparttable}
\usepackage{xspace}
\usepackage{csquotes}
\usepackage{changes}
\usepackage{subfig}
\theoremstyle{thmstyleone}%
\theoremstyle{thmstyletwo}%

\theoremstyle{thmstylethree}%

\def\figref#1{Fig.~\ref{#1}}
\def\tabref#1{Tab.~\ref{#1}}
\def\eqref#1{Eq.~(\ref{#1})}

\newcommand\etal{\emph{et al}\xspace}
\newcommand\ie{i.e.,\xspace}
\newcommand\eg{e.g.,\xspace}

\def\zmp{ZMP\xspace}
\def\dof{DoF\xspace}
\def\map{MAP\xspace}
\def\lipm{LIPM\xspace}
\def\com{CoM\xspace}
\def\imu{IMU\xspace}
\def\ft{F/T\xspace}
\def\ekf{EKF\xspace}
\def\slam{SLAM\xspace}
\def\hri{HRI\xspace}
\def\rgb{RGB\xspace}
\def\rgbd{RGB-D\xspace}
\def\nao{Nao\xspace}

\def\lidar{LiDAR\xspace}

\definechangesauthor{Arindam}
\definechangesauthor{Shahram}
\definechangesauthor{Shubham}
\definechangesauthor{Maren}

\begin{document}

\title[]{Perception for Humanoid Robots}


\author*[1]{\fnm{Arindam} \sur{Roychoudhury}}\email{roychoud@cs.uni-bonn.de}

\author[1]{\fnm{Shahram} \sur{Khorshidi}}\email{khorshidi@cs.uni-bonn.de}

\author[1]{\fnm{Subham} \sur{Agrawal}}\email{sagrawal@cs.uni-bonn.de}

\author[1,2]{\fnm{Maren} \sur{Bennewitz}}\email{maren@cs.uni-bonn.de}

\affil*[1]{\orgdiv{Humanoid Robots Lab}, \orgname{University of Bonn}, \orgaddress{\country{Germany}}}
\affil[2]{\orgname{Lamarr Institute for Machine Learning and Artificial Intelligence}, \orgaddress{\country{Germany}}}


%\abstract{The abstract serves both as a general introduction to the topic and as a brief, non-technical summary of the main results and their implications. Authors are advised to check the author instructions for the journal they are submitting to for word limits and if structural elements like subheadings, citations, or equations are permitted.}

%%================================%%
%% Sample for structured abstract %%
%%================================%%

\abstract{
	\textbf{Purpose of Review} The field of humanoid robotics, perception plays a fundamental role in enabling robots to interact seamlessly with humans and their surroundings, leading to improved safety, efficiency, and user experience. This scientific study investigates various perception modalities and techniques employed in humanoid robots, including visual, auditory, and tactile sensing by exploring recent state-of-the-art approaches for perceiving and understanding the internal state, the environment, objects, and human activities. 
	
	\textbf{Recent Findings} Internal state estimation makes extensive use of Bayesian filtering methods and optimization techniques based on maximum a-posteriori formulation by utilizing proprioceptive sensing. In the area of external environment understanding, with an emphasis on robustness and adaptability to dynamic, unforeseen environmental changes, the new slew of research discussed in this study have focused largely on multi-sensor fusion and machine learning in contrast to the use of hand-crafted, rule-based systems. Human robot interaction methods have  established the importance of contextual information representation and memory for understanding human intentions.

	\textbf{Summary} This review summarizes the recent developments and trends in the field of perception in humanoid robots. Three main areas of application are identified, namely, internal state estimation, external environment estimation, and human robot interaction. The applications of diverse sensor modalities in each of these areas are considered and recent significant works are discussed.
}

\keywords{Humanoid Robots, Perception, Survey, Navigation, State Estimation, Human Robot Interaction}



\maketitle

\section{Introduction}
\label{sec:introduction}

%The Introduction section, of referenced text \cite{bib1} expands on the background of the work (some overlap with the Abstract is acceptable). The introduction should not include subheadings.
%
%Springer Nature does not impose a strict layout as standard however authors are advised to check the individual requirements for the journal they are planning to submit to as there may be journal-level preferences. When preparing your text please also be aware that some stylistic choices are not supported in full text XML (publication version), including coloured font. These will not be replicated in the typeset article if it is accepted. 

\begin{figure*}[ht]
	\centering
	\includegraphics[width=\textwidth]{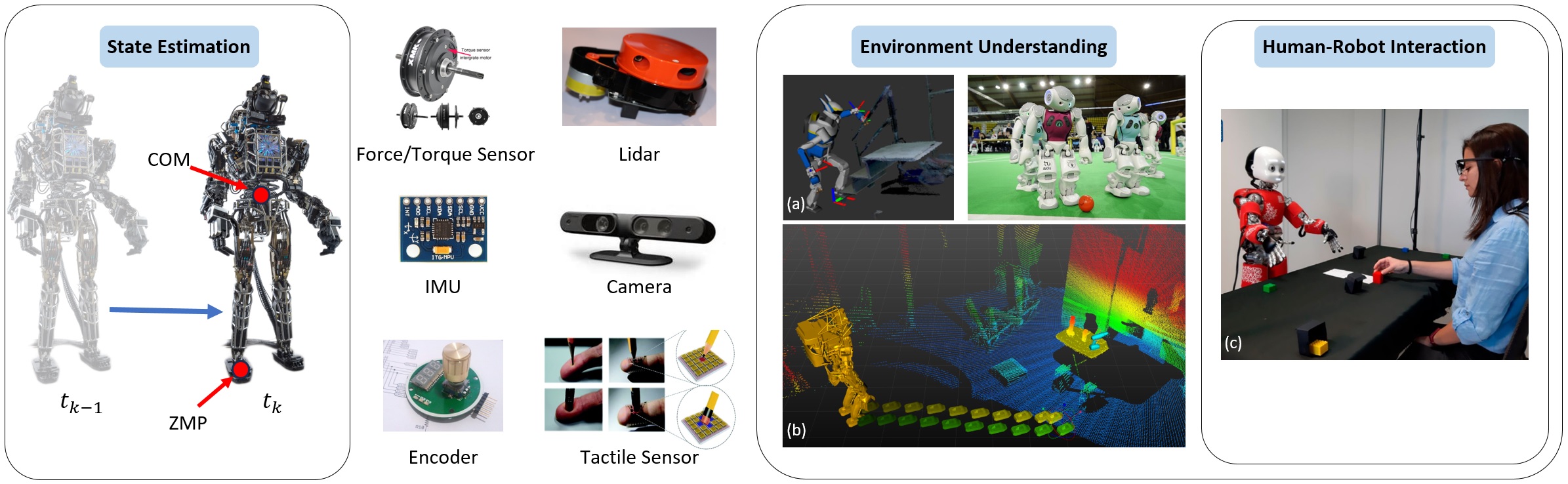}
	\caption{Perception for humanoid robots split into three principal areas. Left: State estimation being used to estimate derived quantities like \com and \zmp from sensors like \imu and joint encoders. Right: Environment understanding has a very broad scope which varies from localization and mapping to environment segmentation for planning and even more application areas. Human Robot Interaction is closely related but deals exclusively with human beings rather than inanimate objects. Center: Few sensors which aid in perception for humanoid robots. Sources for labeled images- (a):\cite{tanguyClosedloopRGBDSLAM2016}, (b): \cite{Fallon14_drift} and (c): \cite{matarese2022perception}.}
	\label{fig:overview}
\end{figure*}

Perception is of paramount importance for robots to establish a model of their internal state as well as the external environment. These models allow the robot to perform its task safely, efficiently and accurately. Perception is facilitated by various types of sensors which gather both proprioceptive and exteroceptive information. Humanoid robots, especially those which are mobile, pose a difficult challenge for the perception process: mounted sensors are susceptible to jerky and unstable motions due to the very high degrees of freedom afforded by the high number of articulable joints present on a humanoid's body, \eg the legs, the hip, the manipulator arms or the neck.

We organize the main areas of perception in humanoid robots into three broad yet overlapping areas for the purposes of this survey, namely, state estimation for balance and joint configurations, environment understanding for navigation, mapping and manipulation, and finally human-robot interaction for successful integration into a shared human workspace, see \figref{fig:overview}. For each area we discuss the popular application areas, the challenges and recent methodologies used to surmount them.

%We see the preference for certain sensor types in some areas. The task of internal state estimation makes overwhelming use of Inertial measurement units (\imu \hspace{-1.5mm}), joint encoders for estimating end effector poses, location of the Zero Moment Point (\zmp), or the base and centroidal state estimation. Sensors such as cameras and lasers are used not only in environment understanding for navigation, localization, mapping, obstacle detection and avoidance \etc but also in human robot interaction. Tactile sensors such as artificial skins and proximity sensors find use in object manipulation and grasping. However these boundaries are nebulous and we see considerable overlap of various sensor modalities in several task areas. 

% High level information about internal state estimation.
Internal state estimation is a critical aspect of autonomous systems, particularly for humanoid robots in order to address both low level stability and dynamics, and as an auxiliary to higher level tasks such as localization, mapping and navigation. Legged robots locomotion is particularly challenging given their inherent under-actuation dynamics and the intermittent contact switching with the ground during motion.

%Internal state estimation is paramount for controlling a humanoid without violating its dynamic constraints which would otherwise cause it to topple during walking or mishandle objects during grasping. In the literature, estimating the \com dynamics and using it as a low dimensional stand-in is the most popular way of wrangling the high dimensionality problem of the full humanoid body. 
%Joint encoders provide accurate readings of the relative positions and orientations of various links relative to the base frame which can be used to have a relatively easy estimate of the position of the \com. However, estimating its velocity, both linear and angular, needs an estimate of the velocity of the base relative to some global inertial frame, which is much more difficult to obtain. This is usually estimated with the help of inertial sensors or \imu and some sort of probabilistic filtering framework such as the Extended Kalman Filter or \ekf to tame the accumulating drift. Use of force torque or \ft sensors fitted to the contact points on a humanoid's feet is also fairly common in \com estimation. In this case, the relationship between the physical quantities of linear momentum and force is used as a basis for estimation. However, the inability of \ft sensors to distinguish between intra-joint shearing forces and external contact forces is a source of considerable error in this approach. 

% High level information about environment undertanding.
The application of external environment understanding has a very broad scope in humanoid robotics but can be roughly divided into navigation and manipulation. Navigation implies the movement of the mobile bipedal base from one location to another without collision thereby leaving the external environment configuration unchanged. On the other hand, manipulation is where the humanoid changes the physical configuration of its environment using its end-effectors.

% High level information about human robot interaction.
It could be argued that human robot interaction or \hri is a subset of environment understanding. However, we have separated the two areas based on their ultimate goals. The goal of environment understanding is to interact with inanimate objects while the goal of \hri is to interact with humans. The set of posed challenges are different though similar principles may be reused. Human detection, gesture and activity recognition, teleoperation, object handover and collaborative actions, and social communications are some of the main areas where perception is used.

% All state estimation related text.
\section{State Estimation}
Recent works on humanoid and legged robots locomotion control have focused extensively on state-feedback approaches \cite{Carpentier21}. Legged robots have highly nonlinear dynamics, and they need high frequency ($1\: kHz$) and low latency ($\textless1\: ms$) feedback in order to have robust and adaptive control systems, thereby adding more complexity to the design and development of reliable estimators for the base and centroidal states, and contact detection.
\subsection{Challenges in State Estimation}
Perceived data is often noisy and biased and it gets magnified in derived quantities. For instance, joint velocities tend to be noisier than joint positions, as these are obtained by numerically differentiating joint encoder values. Rotella \etal \cite{Rotella16-inertial} developed a method to determine joint velocities and acceleration of a humanoid robot using link-mounted Inertial Measurement Units ({\imu}s), resulting in less noise and delay compared to filtered velocities from numerical differentiation.
An effective approach to mitigate biased \imu measurements is to explicitly introduce these biases as estimated states in the estimation framework \cite{Bloesch13}, \cite{Rotella14_state}.

The high dimensionality of humanoids make it computationally expensive to formulate a single filter for the entire state. As an alternative, Xinjilefu \etal \cite{Xinjilefu14_decoupled} proposed decoupling the full state into several independent state vectors, and used separate filters to estimate the pelvis state and joint dynamics.

To account for kinematic modeling errors such as joint backlash and link flexibility, Xinjilefu \etal \cite{Xinjilefu15_com} introduced a method using a Linear Inverted Pendulum Model (\lipm) with an offset which represented the modeling error in the Center of Mass (\com) position and/or external forces. Bae \etal \cite{Bae18_com_compliant} proposed a \com kinematics estimator by including a spring and damper in the \lipm to compensate for modeling errors. To address the issue of link flexibility in the humanoid exoskeleton \emph{Atalante}, Vigne \etal \cite{Vigne20_flexibility} decomposed the full state estimation problem into several independent attitude estimation problems, each corresponding to a given flexibility and a specific \imu relying only on dependable and easily accessible geometric parameters of the system, rather than the dynamic model.

%% ------------------------- LC and TC Estimators --(Deleted) ----------------------------------- 
%\comment[id=Arindam]{Does not talk about what's being perceived. Is this necessary?}
%\deleted{Another challenge in state estimation is the degree of coupling between the position and orientation estimates. Flayols \etal \cite{Flayols17_simple} presented a two-staged \emph{loosely-coupled} or \emph{LC} estimator that comprised of a complementary filter and either a simple Kalman filter or a two-stage weighting algorithm. Both estimators were fast and yielded accurate results on balance and walking. On the other hand, \emph{tightly-couple}d or \emph{TC} approaches were presented in \cite{Bloesch13} \cite{Rotella14_state} and \cite{Fallon14_drift}, which favored a joint estimation of the base states of legged and humanoid robots.} In general, \emph{TC} methods offer several advantages, such as maximizing system observability, enabling online debiasing, and facilitating integration with other sensor modalities \cite{Fourmy22}.

In the remainder of this section, we classify the recent related works on state estimation into three main categories \cite{Camurri20Pronto}: proprioceptive state estimation, which primarily involves filtering methods that fuse high-frequency proprioceptive sensor data; multi-sensor fusion filtering, which integrates exteroceptive sensor modalities into the filtering process; multi-sensor fusion with state smoothing, which employs advanced techniques that leverage the entire history of sensor measurements to refine estimated states. 

Finally, we present a list of available open-source software for state estimation from reviewed literature in \tabref{table_software}.

\begin{table*}[htb]
	\begin{tabular*}{\textwidth}{@{\extracolsep\fill}|c|c|c|c|}
		\toprule
		\thead{\textbf{Paper}} &\thead{\textbf{Software}} & \thead{\textbf{Language}} & \thead{\textbf{Description}} \\
		\midrule
		\multicolumn{4}{|@{}c@{}|}{\thead{\textbf{Extended Kalman Filtering}}}\\ %
		\midrule
		\thead{\cite{Piperakis18_nonlinear}} & \thead{SEROW\cite{Serow18}} & \thead{C++} &  \thead{Multi-sensor state estimation (\imu, joint encoders, visual odometry)}\\
		\midrule
		\thead{\cite{Camurri20Pronto}} & \thead{PRONTO\cite{Pronto}} & \thead{C++} & \thead{Multi-sensor state estimation (\imu, joint encoders, \lidar, camera)}\\
		\midrule
		\thead{\cite{Hartley20_contact}} & \thead{InEKF\cite{InEKF}} & \thead{C++} & \thead{Invariant EKF (using \imu motion model, with different measurement models)} \\
		\midrule
		\multicolumn{4}{|@{}c@{}|}{\thead{\textbf{Factor Graph}}}\\ %
		\midrule
		\thead{\cite{Sola22_wolf}} & \thead{WOLF\cite{WOLF}} & \thead{C++} & \thead{Multi-sensor state smoothing (\imu, joint encoders, \lidar, camera)} \\
		\midrule
		\multicolumn{4}{|@{}c@{}|}{\thead{\textbf{Learning}}}\\ %
		\midrule
		\thead{\cite{Piperakis19_unsupervised}} & \thead{GEM\cite{GEM}} & \thead{Python} &  \thead{Unsupervised gait-phase estimation (\imu, joint encoders, \ft sensors)} \\
		\botrule
	\end{tabular*}
	\caption{Open-source software for humanoid robot state estimation. All cited software are available as ROS packages.}\label{table_software}
\end{table*}

\subsection{Proprioceptive State Estimation}
Proprioceptive sensors provide measurements of the robot's internal state. They are commonly used to compute leg odometry, which captures the drifting pose. For a comprehensive review of the evolution of proprioceptive filters on leg odometry, refer to \cite{Bloesch17}, and \cite{Camurri17}.
\subsubsection{Base State Estimation}
In humanoid robots, the focus is on estimating the position, velocity, and orientation of the \textquote{base} frame, typically located at the pelvis. Recent state estimation approaches in this field often fuse \imu and leg odometry. 

The work by Bloesch \etal \cite{Bloesch13} was a decisive step in introducing a base state estimator for legged robots using a quaternion-based Extended Kalman Filter (\ekf) approach. This method made no assumptions about the robot's gait and number of legs or the terrain structure and included absolute positions of the feet contact points, and \imu bias terms in the estimated states. Rotella \etal \cite{Rotella14_state} extended it to humanoid platforms by considering the full foot plate and adding foot orientation to the state vector. Both works showed that as long as at least one foot remains in contact with the ground, the base absolute velocity, roll and pitch angles, and \imu biases are observable. There are also other formulations for the base state estimation using only proprioceptive sensing in \cite{Hartley20_contact}, \cite{Flayols17_simple}, and \cite{Xinjilefu14_QP}.

%The theory of invariant observer design, investigated by Barrau \etal \cite{Barrau17}, is based on the estimation error being invariant under the action of a matrix Lie group, leading to an error evolution independent of the state trajectory. Hartley \etal \cite{Hartley20_contact} developed an invariant \ekf (\emph{InEKF}) based on the theory of Lie groups. Their results demonstrated improved convergence and consistency of \emph{InEKF} compared to quaternion-based \ekf. Ramadoss \etal \cite{Ramadoss21} employed a similar approach, but a different choice of uncertainty parameterization within the \ekf framework. They also provided a comparison between state-of-the-art flat-foot filters based on the representation choice of state, matrix Lie group error, and system dynamics \cite{Ramadoss22_comparison}. \deleted[id=Arindam,comment={Is it possible to discuss optimization methods w.r.t perception explicitly?}]{Although filtering-based methods are simpler and computationally more efficient, recent advancements in computation power have enabled the effective utilization of optimization-based methods in state estimation. For instance, Xinjilefu \etal \cite{Xinjilefu14_QP} formulated the humanoid state estimation as a Quadratic Programming (\qp) problem, offering several advantages over a nonlinear Kalman filter. \qp does not require the dynamic system to be written in the state space form and can handle equality and inequality constraints in the problem formulation.}

\subsubsection{Centroidal State Estimation}
Centroidal states in humanoid robots include the \com position, linear and angular momentum, and their derivatives. The \com serves as a vital control variable for stability and robust humanoid locomotion, making accurate estimation of centroidal states crucial in control system design for humanoid robots.

When the full 6-axis contact wrench is not directly available to the estimator, \eg the robot gauge sensors measure only the contact normal force, some works have utilized simplified models of dynamics, such as the \lipm \cite{Fourmy22}.

Piperakis \etal \cite{Piperakis16_zmp} presented an \ekf to estimate centroidal variables by fusing joint encoders, \imu, foot sensitive resistors, and later including visual odometry in \cite{Piperakis18_nonlinear}. They formulated the estimator based on the non-linear Zero Moment Point (\zmp) dynamics, which captured the coupling between dynamics behavior in the frontal and lateral planes. Their results showed better performance over Kalman filter formulation based on the \lipm.

Mori \etal \cite{Mori18_com_identification} proposed a centroidal state estimation framework for a humanoid robot based on real-time inertial parameter identification, using only the robot’s proprioceptive sensors (\imu, foot Force/Torque (\ft) sensors, and joint encoders), and the sequential least squares method. They conducted successful experiments deliberately altering the robot's mass properties to demonstrate the robustness of their framework against dynamic inertia changes.

By having 6-axis \ft sensors on the feet, Rotella \etal \cite{Rotella15_momentum} utilized momentum dynamics of the robot to estimate the centroidal quantities. Their nonlinear observability analysis demonstrated the observability of either biases or external wrench. In a different approach, Carpentier \etal \cite{Carpentier16_Spectral} proposed a frequency analysis of the information sources utilized in estimating the \com position, and later for \com acceleration and the derivative of angular momentum \cite{Bailly19_com_derivative}. They introduced a complementary filtering technique that fuses various measurements, including \zmp position, sensed contact forces, and geometry-based reconstruction of the \emph{CoM} by using joint encoders, according to their reliability in the respective spectral bandwidth.

%\deleted[id=Arindam,comment={Is it possible to fuse these citations if they don't introduce anything new to do with the \imu or other sensors?}]{Bailly \etal \cite{Bailly19_com_derivative} extended this method by incorporating \com acceleration and the derivative of angular momentum into the estimated variables, resulting in more accurate estimation of \com position compared to a simple \ekf . \cite{Bailly21_DDP} formulated the estimation of centroidal dynamics as a maximum a-posteriori (\emph{MAP}) problem, and used a differential dynamic programming approach to solve it, resulting in better performance compared to \ekf and complementary filtering estimators.}
%

\begin{figure*}[htb]
	\centering
	\includegraphics[width=\linewidth]{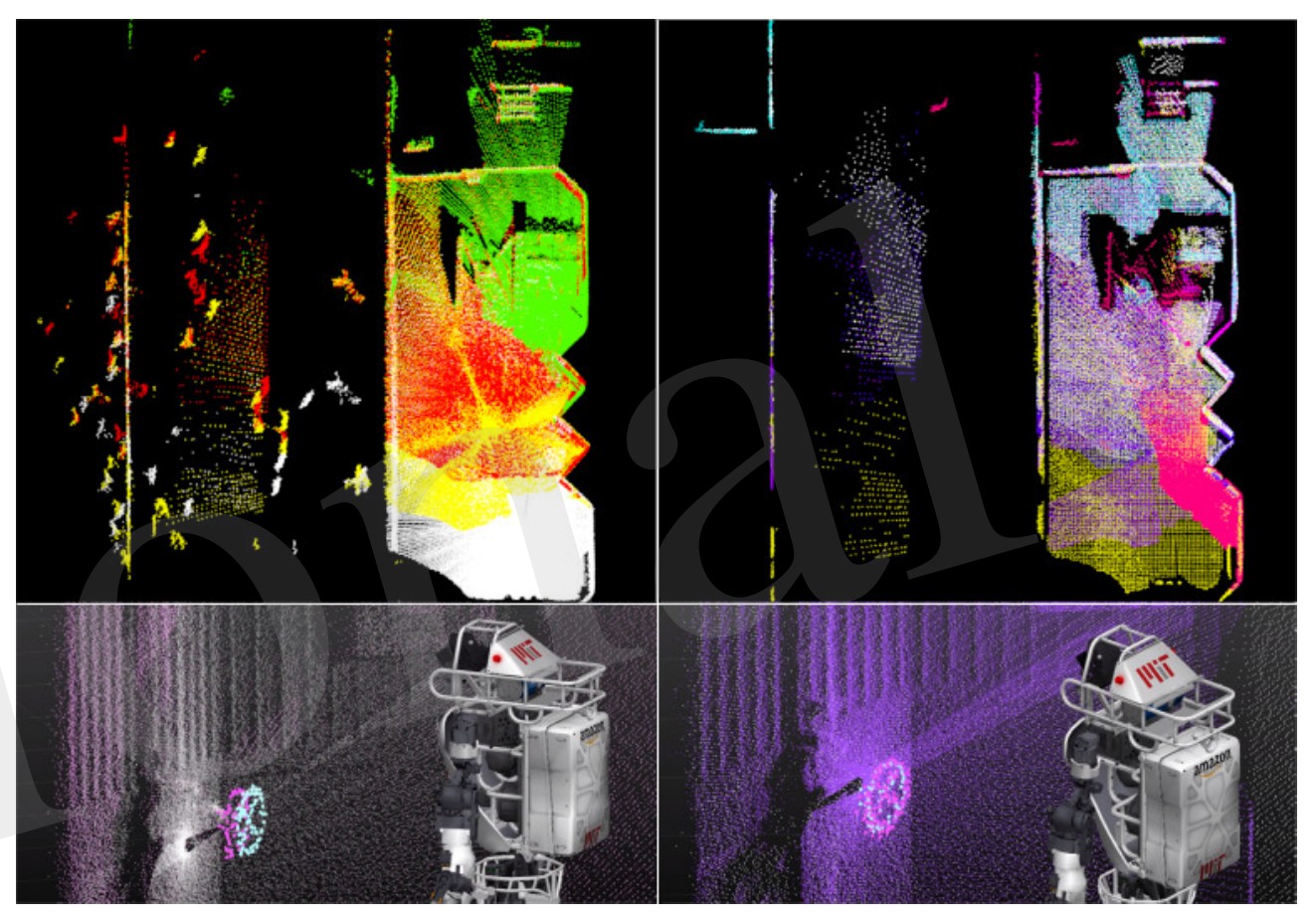}
	\caption{State estimation with multi-sensor filtering, integrating \lidar for drift correction and localization. Top row, filtering people from raw point cloud. Bottom row, state estimation and localization with iterative closest point correction on filtered point cloud. From \cite{Camurri20Pronto}.}
	\label{fig:camurri}
\end{figure*}

\subsubsection{Contact Detection and Estimation}
Feet contact detection plays a crucial role in locomotion control, gait planning, and proprioceptive state estimation in humanoid robots. Recent approaches can be categorized into two main groups: those directly utilizing measured ground reaction wrenches, and methods integrating kinematics and dynamics to infer the contact status by estimating the ground reaction forces. Fallon \etal \cite{Fallon14_drift} employed a Schmitt trigger with a 3-axis foot \ft sensor to classify contact forces and used a simple state machine to determine the most reliable foot for kinematic measurements. Piperakis \etal \cite{Piperakis18_nonlinear} adapted a similar approach by utilizing pressure sensors on the foot.

Rotella \etal \cite{Rotella18_unsupervised} presented an unsupervised method for estimating contact states by using fuzzy clustering on only proprioceptive sensor data (foot \ft and \imu sensing), surpassing traditional approaches based on measured normal force. By including the joint encoders in proprioceptive sensing, Piperakis \etal \cite{Piperakis19_unsupervised} proposed an unsupervised learning framework for gait phase estimation, achieving effectiveness on uneven/rough terrain walking gaits. They also developed a deep learning framework by utilizing \ft and \imu sensing in each leg, to determine the contact state probabilities \cite{Piperakis22_robust}. The generalizability and accuracy of their approach was demonstrated on different robotic platforms. Furthermore, Maravgakis \etal \cite{Maravgakis23_probabilistic} introduced a probabilistic contact detection model, using only \imu sensors mounted on the end effector. Their approach estimated the contact state of the feet without requiring training data or ground truth labels.

Another active research field in humanoid robots is monitoring and identifying contact points on the robot's body. Common approaches focus on proprioceptive sensing for contact localization and identification. Flacco \etal \cite{Flacco16_residual} proposed using an internal residual of external momentum to isolate and identify singular contacts, along with detecting additional contacts with known locations. Manuelli \etal \cite{Manuelli16_externalforce} introduced a contact particle filter for detecting and localizing external contacts, by only using proprioceptive sensing, such as 6-axis \ft sensors, capable of handling up to 3 contacts efficiently. Vorndamme \etal \cite{Vorndamme21_Sami_multicontact} developed a real-time method for multi-contact detection using 6-axis \ft sensors distributed along the kinematic chain, capable of handling up to 5 contacts. 
Vezzani \etal \cite{vezzaniMemoryUnscentedParticle2017} proposed a memory unscented particle filter algorithm for real-time 6 Degrees of freedom (\dof) tactile localization using contact point measurements made by tactile sensors.

\subsection{Multi-Sensor Fusion Filtering}
One drawback of base state estimation using proprioceptive sensing is the accumulation of drift over the time, due to sensor noise. This drift is not acceptable for controlling highly dynamic motions, therefore it is typically compensated by integrating other sensor modalities from exteroceptive sensors, such as cameras, depth cameras, and \lidar.

Fallon \etal \cite{Fallon14_drift} proposed a drift-free base pose estimation method by incorporating \lidar sensing into a high-rate EKF estimator using a Gaussian particle filter for laser scan localization. Although their framework eliminated the drift, a pre-generated map was required as input. Piperakis \etal \cite{Piperakis19_outlier} introduced a robust Gaussian \ekf to handle outlier detection in visual/\lidar measurements for humanoid walking in dynamic environments. To address state estimation challenges in real-world scenarios, Camurri \etal \cite{Camurri20Pronto} presented Pronto, a modular open-source state estimation framework for legged robots \figref{fig:camurri}. It combined proprioceptive and exteroceptive sensing, such as stereo vision and \lidar, using a loosely-coupled EKF approach.

\subsection{Multi-Sensor Fusion with State Smoothing}
So far, we have explored filtering methods based on Bayesian filtering for sensor fusion and state estimation. However, as the number of states and measurements increases, computational complexity becomes a limitation. Recent advancements in computing power and nonlinear solvers have popularized non-linear iterative maximum a-posteriori (\map) optimization techniques, such as factor graph optimization.

To address the issue of visual tracking loss in visual factor graphs, Hartley \etal \cite{Hartley18_legged} introduced a factor graph framework that integrated forward kinematic and pre-integrated contact factors. The work was extended by incorporating the influence of contact switches and associated uncertainties \cite{Hartley18_Hybrid}. Both works showed that the fusion of contact information with \imu and vision data provides a reliable odometry system for legged robots.

Sola \etal \cite{Sola22_wolf} presented an open-source modular estimation framework for mobile robots based on factor graphs. Their approach offered systematic methods to handle the complexities arising from multi-sensory systems with asynchronous and different-frequency data sources. This framework was evaluated on state estimation for legged robots and landmark-based visual-inertial \slam for humanoids by Fourmy \etal \cite{Fourmy22}.

% All environment understanding related text.
\section{Environment Understanding} 
\label{sec:environment-understanding}

Environment understanding is a critical area of research for humanoid robots, enabling them to effectively navigate through and interact with complex and dynamic environments. This field can be broadly classified into two key categories: 1. localization, navigation and planning for the mobile base, and 2. object manipulation and grasping. 

%To accomplish these tasks, humanoid robots rely on a diverse range of sensors. Cameras provide visual perception, capturing images and videos to enable scene understanding and object recognition. Laser, \lidar and \rgbd cameras provide depth information and enable environment mapping, obstacle detection, and collision avoidance. {\imu}s incorporate accelerometers and gyroscopes to measure the robot's motion and orientation in three-dimensional space, aiding in localization and motion planning. Joint encoders capture the precise positioning and movement of robot joints, allowing for accurate control and coordination of robot movements. A plethora of other sensors are used as well, each catering to its own niche.

\subsection{Perception in Localization, Navigation and Planning}
\label{subsec:navigation-planning-localization}

Localization focuses on precisely and continuously estimating the robot's position and orientation relative to its environment. 
Planning and navigation involve generating optimal paths and trajectories for the robot to reach its desired destination while avoiding obstacles and considering task-specific constraints. 

\subsubsection{Localization, Mapping and \slam}
\label{subsubsec:visual-slam-localization}

Localization and \slam (simultaneous localization and mapping) relies primarily on visual sensors such as cameras and lasers but often additionally use encoders and {\imu}s to enhance estimation accuracy. 

\paragraph{Localization}
\label{para:localization}

Indoor environments are usually considered structured, characterized by the presence of well-defined, repeatable and often geometrically consistent objects. Landmarks can be uniquely identified by encoded vectors obtained from visual sensors such as depth or \rgb cameras allowing the robot to essentially build up a visual map of the environment and then compare newly observed landmarks against a database to localize via object or landmark identification. In recent years, the use of handcrafted image features such as  SIFT and SURF and feature dictionaries such as the Bag-of-Words (BoW) model in landmark representation has been superseded by feature representations \emph{learned} through training on large example sets, usually by variants of artificial neural networks such as convolutional neural networks (CNNs). CNNs have also outperformed classifiers such as support vector machines (SVMs) in deriving inferences \cite{wozniakSceneRecognitionIndoor2018} \cite{wozniakPlaceInferenceGraphBased2021}. However, several rapidly evolving CNN architectures exist. Ovalle-magallanes \etal \cite{ovalle-magallanesTransferLearningHumanoid2021} performed a comparative study of four such networks while successfully localizing in a visual map.

The \emph{RoboCup Soccer League} is popular in humanoid research due to the visual identification and localization challenges it presents. \cite{speckRealTimeBallLocalization2019}, \cite{teimouriRealTimeBallDetection2019} and \cite{gabelJetsonWhereBall2019} are some examples of real-time, CNN based ball detection approaches utilizing \rgb cameras developed specifically for RoboCup. Cruz \etal \cite{cruzDeepLearningApplied2021} could additionally estimate player poses, goal locations and other key pitch features using intensity images alone. Due to the low on-board computational power of the humanoids, others have used fast, low power external mobile GPU boards such as the Nvidia Jetson to aid inference \cite{gabelJetsonWhereBall2019} \cite{chatterjeeRealTimeObjectDetection2020}.

Unstructured and semi-structured environments are encountered outdoors or in hazardous and disaster rescue scenarios. They have a dearth of reliably trackable features, unpredictable lighting conditions and are challenging for gathering training data. Thus, instead of features, researchers have focused on raw point clouds or combining different sensor modalities for navigating such environments.
Starr \etal \cite{starrEvidentialSensorFusion2017} presented a sensor fusion approach which combined long-wavelength infrared stereo vision and a spinning \lidar for accurate rangefinding in smoke-obscured environments. 
Nobili \etal \cite{nobiliOverlapbasedICPTuning2017a} successfully localized robots constrained by a limited field-of-view \lidar in a semi-structured environment. They proposed a novel strategy for tuning outlier filtering based on point cloud overlap which achieved good localization results in the DARPA Robotics Challenge Finals.
Raghavan \etal \cite{raghavanStudyLowDriftState2018} presented simultaneous odometry and mapping by fusing \lidar and kinematic-inertial data from \imu, joint encoders, and foot \ft sensors while navigating a disaster environment.

\paragraph{\slam}
\label{para:appearance-based-slam}

\slam subsumes localization by the additional map construction and loop closing aspects, whereby the robot has to re-identify and match a place which was visited sometime in the past, to its current surroundings and adjust its pose history and recorded landmark locations accordingly.
A humanoid robot which is intended to share human workspaces needs to deal with moving objects, both rapid and slow, which could disrupt its mapping and localizing capabilities. Thus, recent works on \slam have focused on handling the presence of dynamic obstacles in visual scenes. While the most popular approach remains sensor fusion \cite{scona2017direct} \cite{tanguy2019closed}, other purely visual approaches have also been proposed, such as,
\cite{zhang2020flowfusion} which introduced a dense \rgbd \slam solution that utilized optical flow residuals to achieve accurate and efficient dynamic/static segmentation for camera tracking and background reconstruction. 
Zhang \etal \cite{zhang2018dense} took a more direct approach which employed deep learning based human detection, and used graph-based segmentation to separate moving humans from the static environment. They further presented a \slam benchmark dedicated to dynamic environment \slam solutions \cite{zhang2019hrpslam}. It included \rgbd data acquired from an on-board camera on the \emph{HRP-4} humanoid robot, along with other sensor data.
Adapting publicly available \slam solutions and tailoring it for humanoid use is not uncommon.
Sewtz \etal \cite{sewtz2021robust} adapted the Orb-Slam \cite{mur-artalORBSLAMVersatileAccurate2015} for a multi-camera setup on the DLR \emph{Rollin' Justin} System while Ginn \etal \cite{ginnMonocularORBSLAMHumanoid2018} did it for the \emph{iGus}, a midsize humanoid platform, to have low computational demands.

\subsubsection{Navigation and Planning}
\label{subsubsec:navigation-planning}

Navigation and planning algorithms use perception information to generate a safe, optimal and reactive path, considering obstacles, terrain, and other constraints. 

\paragraph{Local Planning}
\label{para:local-planning}

Local planning or reactive navigation is generally concerned with local real-time decision-making and control, allowing the robot to actively respond to perceived changes in the environment and adjust its movements accordingly. Especially in highly controlled applications rule-based, perception driven navigation is still popular and yields state-of-the-art performance both in terms of time demands and task accomplishment.
Bista \etal \cite{bistaCombiningLineSegments2017} achieved real-time navigation in indoor environments by representing the environment by key \rgb images, and deriving a control law based on common line segments and feature points between the current image and nearby key images. 
Regier \etal \cite{regierClassifyingObstaclesExploiting2018} determined appropriate actions based on a pre-defined set of mappings between object class and action. A CNN was used to classify objects from monocular \rgb vision. 
Ferro \etal \cite{ferroVisionBasedNavigationOmnidirectional2019} integrated information from a monocular camera, joint encoders, and an \imu to generate a collision-free visual servo control scheme.
Juang \etal \cite{juangRobustVisualLinefollowing2020} developed a line follower which was able to infer forward, lateral and angular velocity commands using path curvature estimation and PID control from monocular \rgb images.
Magassouba \etal \cite{magassoubaAuralServoSensorBased2018} introduced an aural servo framework based on auditory perception, enabling robot motions to be directly linked to low-level auditory features through a feedback loop.

We also see the use of a diverse array of classifiers to learn navigation schemes from perception information. Their generalization capability allows adaptation to unforeseen obstacles and events in the environment.
Abiyev \etal \cite{abiyevRobotPathfindingUsing2017} presented a vision-based path-finding algorithm which segregated captured images into free and occupied areas using an SVM. 
Lobos-tsunekawa \etal \cite{lobos-tsunekawaVisualNavigationBiped2018} and Silva \etal \cite{silvaRoboticCognitionMeans2021} proposed deep learned visual (\rgb) navigation systems for humanoid robots which were able to achieve real time performance. The former used a reinforcement learning (RL) system with an actor-critic architecture while the latter utilized a decision tree of deep neural networks deployed on a soccer playing robot.

\paragraph{Global Planning}
\label{para:planning}

These algorithms operate globally, taking into account long-term objectives and optimize movements to minimize costs, maximize efficiency, or achieve a specific outcome on the basis of a perceived environment model.

Footstep Planning is a crucial part of humanoid locomotion and has generated substantial research interest for itself. Recent works exhibit two primary trends related to perception. The first is providing humanoids the capability of rapidly perceiving changes in the environment and reacting through fast re-planning. The second endeavors to segment and/or classify uneven terrains to find stable 6 \dof footholds for highly versatile navigation.

\begin{figure}[tb]
	\centering
	\includegraphics[width=\linewidth]{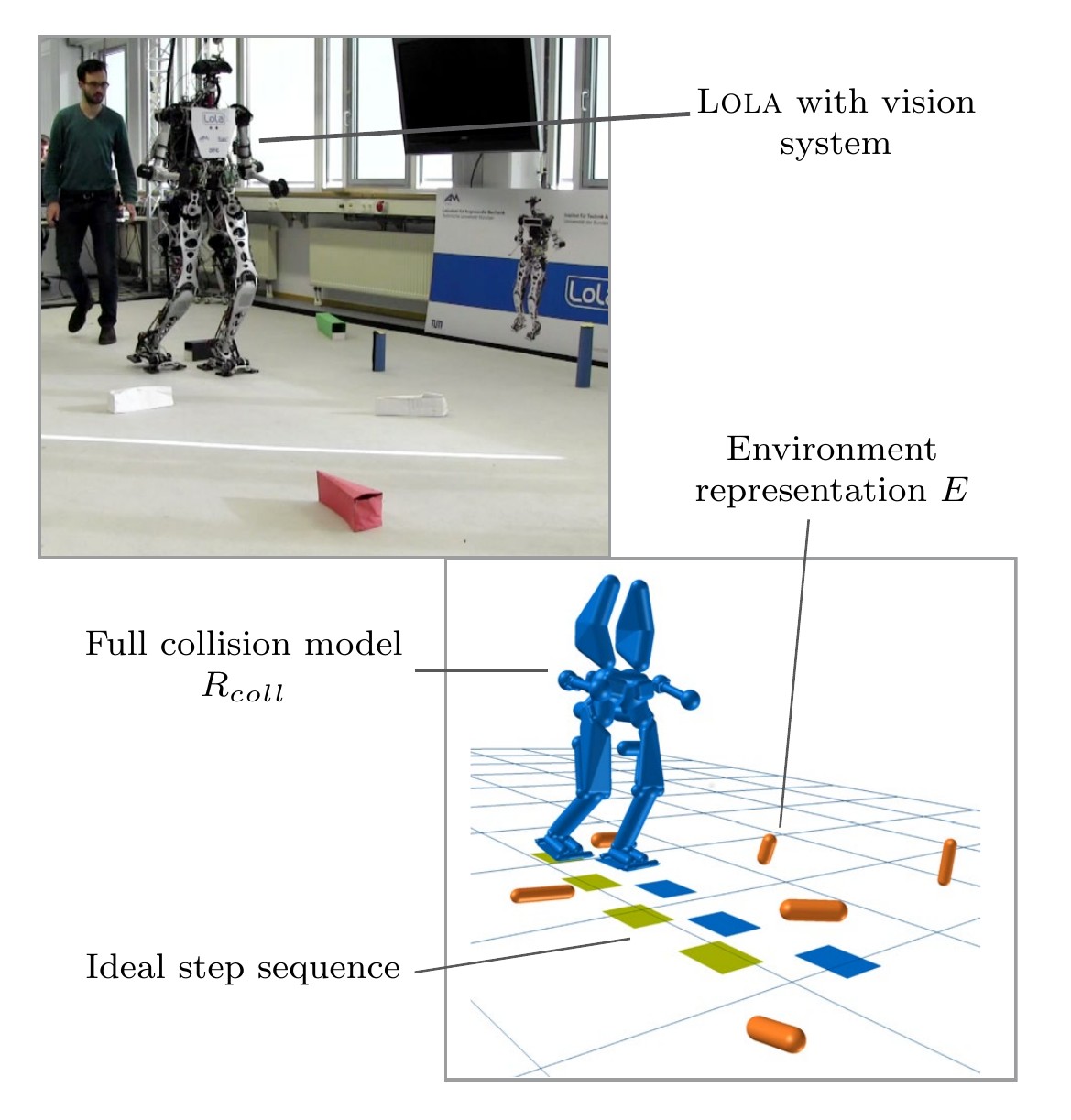}
	\caption{Footstep planning on the humanoid \emph{Lola} from \cite{hildebrandtVersatileRobustBipedal2019}. Top left: The robot's vision system and a human causing disturbance. Bottom right: The collision model with geometric obstacle approximations.}
	\label{fig:hildebrandt}
\end{figure}

Tanguy \etal \cite{tanguy2019closed} proposed a model predictive control (MPC) scheme that fused visual \slam and proprioceptive \ft sensors for accurate state estimation. This allowed rapid reaction to external disturbances by adaptive stepping leading to balance recovery and improved localization accuracy.
Hildebrandt \etal \cite{hildebrandtVersatileRobustBipedal2019} used the point cloud from an \rgbd camera to model obstacles as swept-sphere-volumes (SSVs) and step-able surfaces as convex polygons for real-time reactive footstep planning with the \emph{Lola} humanoid robot. Their system was capable of handling rough terrain as well as external disturbances such as pushes (see \figref{fig:hildebrandt}).
Others have also used geometric primitives to aid in footstep planning, such as surface patches for foothold representation \cite{kanoulasFootstepPlanningRough2018} \cite{kanoulasCurvedPatchMapping2019}, environment segmentation to find step-able regions, such as 2D plane segments embedded in 3D space \cite{bertrandDetectingUsablePlanar2020} \cite{roychoudhury3DPolygonalMapping2022}, or represented obstacles by their polygonal ground projections \cite{missuraPolygonalPerceptionMobile2020}.
Suryamurthy \etal \cite{suryamurthyTerrainSegmentationRoughness2019} assigned pixel-wise terrain labels and rugosity measures using a CNN consuming \rgb images for footstep planning on a \emph{CENTAURO} robot.
%\cite{yangPredictingPolarizationSemantics2018b} used deep architectures for predicting per-pixel polarization information from monocular RGB images, specifically focusing on challenging surfaces such as water hazards, transparent glasses, and metallic surfaces. This was achieved by attaching polarizers on low cost RGB cameras instead of using expensive micro-polarizer cameras. The method was experimentally validated on a wearable exoskeleton humanoid robot, demonstrating accurate results.

%\textbf{Viewpoint Planning} focuses on determining the optimal location, orientation, and trajectory of the robot's sensors or cameras to gather relevant information about the environment and objects of interest, taking into account factors such as task requirements, sensor capabilities, occlusion constraints, and the desired field of view. 

Whole Body Planning in humanoid robots involves the coordinated planning and control of the robot's entire body to achieve an objective.
Coverage planning is a subset of whole body planning where a minimal sequence of whole body robot poses are estimated to completely explore a 3D space via robot mounted visual sensors \cite{osswaldEfficientCoverage3D2017} \cite{osswaldGPUAcceleratedNextBestViewCoverage2018}. Target finding is a special case of coverage planning where the exploration stops when the target is found \cite{monicaHumanoidRobotNext2019} \cite{tsuruOnlineObjectSearching2021}. These concepts are related primarily to view planning in computer vision. 
%Osswald \etal \cite{osswaldEfficientCoverage3D2017} \cite{osswaldGPUAcceleratedNextBestViewCoverage2018} used inverse reachability maps for covering known static and articulated 3D environments using a robot with a monocular camera, whereas Monica \etal \cite{monicaHumanoidRobotNext2019} and Tsuru \etal \cite{tsuruOnlineObjectSearching2021} used off the shelf \slam solutions to simultaneously explore and map a 3D environment with the aim of discovering the target object.
%\cite{liuTargetRecognitionHeavy2018a} studied target recognition and heavy load trolley control in pushing operations. A monocular vision ranging method was used for searching and locating the target and posture closed-loop was used for the control of a heavily loaded robot. 
%\cite{kumagaiPerceptionBasedLocomotion2018b} presented a perception-based locomotion system for humanoids that could adapt to occlusion and task constraints. They combined visual odometry with kinematics for localization and environmental mapping and developed an approach for correcting locomotion errors by locally compensating footsteps and performing an online assessment of the need for global re-planning. The system utilized ground aligned obstacle point clouds obtained from laser scans to enable accurate target reaching and improved locomotion abilities.
%Jin \etal \cite{jinEnhancingBinocularDepth2018} proposed a perception and action based cyclic RL framework which alternatively enhanced stereo based depth estimation and performed behavioral tasks by integrating sensory-invariance driven action with object-size invariance.
In other applications, Wang \etal \cite{wangFormationBuildingCollision2019} presented a method for trajectory planning and formation building of a robot fleet using local positions estimated from onboard optical sensors and Liu \etal \cite{liuTemporalPlanningBasedChoreography2023} presented a temporal planning approach for choreographing dancing robots in response to microphone-sensed music.
%\cite{alatiAnticipatingNextGoal2020} used deep learning to infer the next goal while a robot was engaged in an activity. \emph{Mask-RCNN} was used to provide image segmentation for initial goal recognition. The model was evaluated in a warehouse environment with a humanoid robot assisting a maintenance technician.

\subsection{Perception in Grasping and Manipulation}
\label{subsec:manipulation-grasping}

Manipulation and grasping in humanoid robots involve their ability to interact with objects of varying shapes, sizes, and weights, to perform dexterous manipulation tasks using their sensor equipped end-effectors which provide visual or tactile feedback for grip adjustment. 

\paragraph{Grasp Planning}
\label{para:grasp-planning}

Grasp planning is a lower level task specifically focused on determining the optimal manipulator pose sequence to securely and effectively grasp an object. Visual information is used to find grasping locations and also as a feedback to optimize the difference between the target grasp pose and the current end-effector pose.

Schmidt \etal \cite{schmidtGraspingUnknownObjects2018} utilized a CNN trained on object depth images and pre-generated analytic grasp plans to synthesize grasp solutions. The solution generated full end-effector poses and could generate poses not limited to the camera view direction. 
Vezzani \etal \cite{vezzaniGraspingApproachBased2017} modeled the shape and volume of the target object captured from stereo vision in real-time using super-quadric functions allowing grasping even when parts of the object were occluded.
Vicente \etal \cite{vicenteMarkerlessVisualServoing2017} and Nguyen \etal \cite{nguyenTransferringVisuomotorLearning2018} focused on achieving accurate hand-eye coordination in humanoids equipped with stereo vision. While the former compensated for kinematic calibration errors between the robot's internal hand model and captured images using particle based optimization, the latter trained a deep neural network predictor to estimate the robot arm's joint configuration.
Nguyen \etal \cite{nguyenObjectbasedAffordancesDetection2017} proposed a combination of CNNs and dense conditional random fields (CRFs) to infer action possibilities on an object (affordances) from \rgb images.

\begin{figure}[tb]
	\centering
	\includegraphics[width=\linewidth]{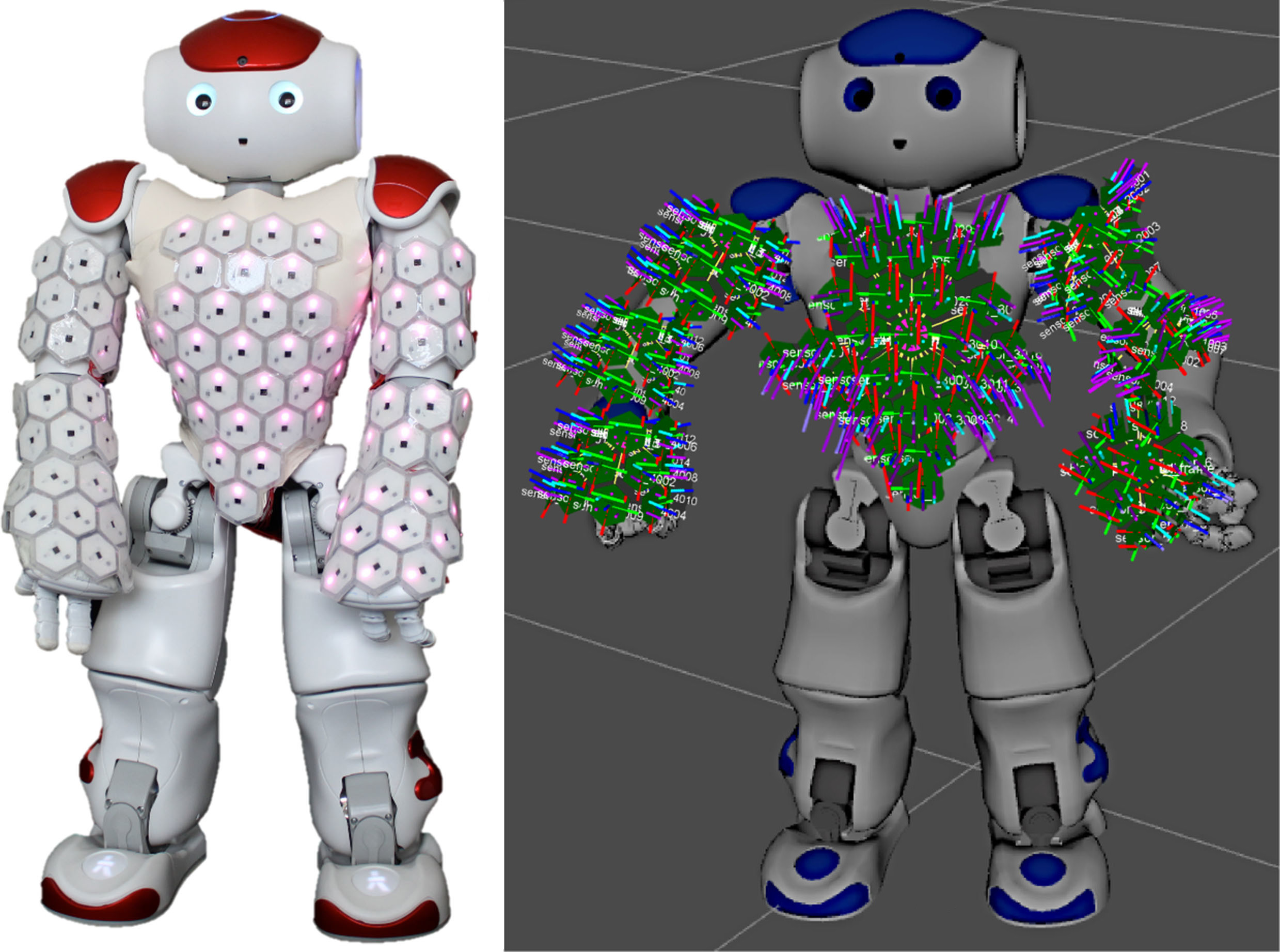}
	\caption{Left: A \nao humanoid equipped with artificial skin cells on the chest, hand, fore arm, and upper arm. Right: Visualization of the skin cell coordinate frames on the \nao. Figure taken from \cite{kaboliRobustTactileDescriptors2018a}.}
	\label{fig:kaboli}
\end{figure}

Tactile sensors, such as pressure-sensitive skins or fingertip sensors, provide feedback about the contact (surface normal) forces, slip detection, object texture, and shape information during object grasping. 
%\textbf{Proximity} sensors, on the other hand, enable the robot to detect the presence and proximity of objects in its environment. 
%Many researchers have focused on developing innovative hardware for the humanoid hand for effective tactile sensing. 
%\cite{paulinoLowcost3axisSoft2017} presented a low-cost 3-axis tactile force sensor based on magnetic and Hall-effect sensor technologies for the human-friendly robot \emph{Vizzy}. 
%\cite{kimSixAxisForceTorque2020} introduced a six-axis \ft fingertip sensor designed for a humanoid robot hand. 
%\cite{paulinoLowcost3axisSoft2017} and \cite{kimSixAxisForceTorque2020} respectively presented 3-axis and 6-axis fingertip sensors for humanoid robot hands. The former combined magnetic and Hall-effect sensing technologies and the latter could also measure shear force using the eccentricity of two cylinders.
%\cite{hoffmannRoboticHomunculusLearning2017} investigated the representation of the skin surface formed from tactile stimulation on a humanoid equipped with artificial pressure-sensitive skin. Representations were formed using self-organizing maps (SOMs).
%which were modified to restrict the maximum receptive field size of neuron groups in response to touch, inspired by biological findings. 
Kaboli \etal \cite{kaboliRobustTactileDescriptors2018a} extracted tactile descriptors for material and object classification agnostic to various sensor types such as dynamic pressure sensors, accelerometers, capacitive sensors, and impedance electrode arrays. A \nao with artificial skin used for their experiments is shown in \figref{fig:kaboli}.
%Tsuji \etal \cite{tsujiProximityContactSensor2020} combined time-of-flight (ToF) sensors for proximity sensing and self-capacitance sensors for tactile sensing respectively, providing measurements with high sensitivity and wide range. 
%The ToF sensor was used to determine the distance between the robot and objects in the proximity range, while the self-capacitance sensor detected objects before and during contact. 
Hundhausen \etal \cite{hundhausenFastReactiveGrasping2021} introduced a soft humanoid hand equipped with in-finger integrated cameras and an in-hand real-time image processing system based on CNNs for fast reactive grasping.

\paragraph{Manipulation Planning}
\label{para:manipulation-planning}

Manipulation planning involves the higher-level decision-making process of determining how the robot should manipulate an object once it is grasped. It generates a sequence of motions or actions which is updated based on the continuously perceived robot and grasped object state. 

Deep recurrent neural networks (RNNs) are capable of predicting the next element in a sequence based on the previous elements. This property is exploited in manipulation planning by breaking down a complex task into a series of manipulation commands generated by RNNs based on past commands. These networks are capable of mapping features extracted from a sequence of \rgb images, usually by CNNs, to a sequence of motion commands \cite{nguyenTranslatingVideosCommands2017a} \cite{kasePutinBoxTaskGenerated2018}.
Inceoglu \etal \cite{inceogluFailureDetectionUsing2018} presented a multimodal failure monitoring and detection system for robots which integrated high-level proprioceptive, auditory, and visual information  during manipulation tasks.
Robot assisted dressing is a challenging manipulation task that has been addressed by multiple authors.
Zhang \etal \cite{zhangPersonalizedRobotassistedDressing2017} utilized a hierarchical multi-task control strategy to adapt the humanoid robot \emph{Baxter}'s applied forces, measured using joint torques, to the user's movements during dressing.
By tracking the subject human's pose in real-time using capacitive proximity sensing with low latency and high signal-to-noise ratio, Erickson \etal \cite{ericksonTrackingHumanPose2018} developed a method to  adapt to human motion and adjust for errors in pose estimation during dressing assistance by the \emph{PR2} robot. 
Zhang \etal \cite{zhangLearningGraspingPoints2020} computed suitable grasping points on garments from depth images using a deep neural network to facilitate robot manipulation in robot-assisted dressing tasks.

\section{Human-Robot Interaction}

Human robot interaction is a subset of environment understanding which deals with interactions with humans as opposed to inanimate objects. In order to achieve this, a robot needs diverse capabilities ranging from detecting humans, recognizing their pose, gesture, and emotions, to predicting their intent and even proactively performing actions to ensure a smooth and seamless interaction. 

There are two main challenges to perception in HRI - perception of users, and inference which involves making sense of the data and making predictions.

\subsection{Perception of Users}
This involves identifying humans in the environment, detecting their pose, facial features, and objects they interact with. This information is crucial for action prediction and emotion recognition \cite{narayanan2020proxemo}. Robots rely on vision-based, audio-based, tactile-based, and range sensor-based sensing techniques for detection as explained in this survey on perception methods of social robots done by \cite{yan2014survey}.

Robinson \etal \cite{robinson2023robotic} showed how vision-based techniques have evolved from using facial features, motion features, and body appearance to deep learning-based approaches. Motion-based features separate moving objects from the background to detect humans. Body appearance-based algorithms use shape, curves, posture, and body parts to detect humans. Deep learning models like R-CNN, Faster R-CNN, and YOLO have also been applied for human detection \cite{yan2014survey}.

Pose detection is essential for understanding human body movements and postures. Sensors such as \rgb cameras, stereo cameras, depth sensors, and motion tracking systems are used to extract pose information. This was explained in detail by M{\"o}ller \etal \cite{moller2021survey} in their survey of human-aware robot navigation. Facial features play a significant role in pose detection as they provide additional points of interest and enable emotion recognition \cite{samadiani2019review}. 
A great demonstration of detecting pose and using it for bi-manual robot control using an \rgbd range sensor was shown by Hwang \etal \cite{hwangRealTimePoseImitation2019}. The system employed a CNN from the \emph{OpenPose} package to extract human skeleton poses, which were then mapped to drive robotic hands. The method was implemented on the \emph{CENTAURO} robot and successfully performed box and lever manipulation tasks in real-time. They presented a real-time pose imitation method for a mid-size humanoid robot equipped with a servo-cradle-head \rgbd vision system. Using eight pre-trained neural networks, the system accurately captured and imitated 3D motions performed by a target human, enabling effective pose imitation and complex motion replication in the robot. Lv \etal \cite{lvGuLiMHybridMotion2022} presented a novel motion synchronization method called \emph{GuLiM} for teleoperation of medical assistive robots, particularly in the context of combating the COVID-19 pandemic. Li \etal \cite{liMobileRobotHandArm2020} presented a multimodal mobile teleoperation system that integrated a vision-based hand pose regression network and an \imu-based arm tracking method. The system allowed real-time control of a robot hand-arm system using depth camera observations and \imu readings from the observed human hand, enabled through the \emph{Transteleop} neural network which generated robot hand poses based on a depth image input of a human hand.

Audio communication is vital for human interaction, and robots aim to mimic this ability. Microphones are used for audio detection, and speakers reproduce sound. Humanoid robots are usually designed to be binaural \ie they have two separate microphones at either side of the head which receive transmitted sound independently. Several researchers have focused on this property to localize both the sound source and the robot in complex auditory environments. Such techniques are used in speaker localization, as well as other semantic understanding tasks such as automatic speech recognition (ASR), auditory scene analysis, emotion recognition, and rhythm recognition \cite{yan2014survey},\cite{badr2020review}.

Benaroya \etal \cite{benaroyaBinauralLocalizationMultiple2018} employed non-negative tensor factorization for binaural localization of multiple sound sources within unknown environments. Schymura \etal \cite{schymuraExtendingLinearDynamical2018} focused on combined audio-visual speaker localization and proposed a closed-form solution to compute dynamic stream weighting between audio and visual streams, improving the state estimation in a reverberant environment. The previous study was extended to incorporate dynamic stream weights into nonlinear dynamical systems which improved speaker localization performance even further \cite{schymuraAudiovisualSpeakerTracking2020}. D{\'a}vila-Chac{\'o}n \etal \cite{davila-chaconEnhancedRobotSpeech2019} used a spiking and a feed-forward neural network for sound source localization and ego noise removal respectively to enhance ASR in challenging environments. 
Trowitzsch \etal \cite{trowitzschJoiningSoundEvent2020} presented a joint solution for sound event identification and localization, utilizing spatial audio stream segregation in a binaural robotic system.

Ahmad \etal \cite{ahmad2022survey} in their survey on physiological signal-based emotion recognition showed that physiological signals from the human body, such as such as heart rate, blood pressure, body temperature, brain activity, and muscle activation can provide insights into emotions.
Tactile interaction is an inherent part of natural interaction between humans and the same holds true for robots interacting with humans as well. The type of touch can be used to infer a lot of things such as the human’s state of mind, the nature of the object, what is expected out of the interaction, etc. \cite{yan2014survey}. Mainly two kinds of tactile sensors are used for this purpose - sensors embedded on the robot’s arms and grippers, and cover based sensors which are used to detect touch across entire regions or the whole body \cite{yan2014survey}. Khurshid \etal \cite{khurshidEffectsGripForceContact2017} investigated the impact of grip-force, contact, and acceleration feedback on human performance in a teleoperated pick-and-place task. Results indicated that grip-force feedback improved stability and delicate control, while contact feedback improved spatial movement but may vary depending on object stiffness.

\subsection{Inference}
An important aspect of inference with all the detected data from the previous section is regarding aligning the perspective of the user and the robot. This allows the robot to better understand the intent of the user regarding the objects or locations they are looking at. This skill is called \emph{perspective taking} and requires the robot to consider and understand other individuals through motivation, disposition, and contextual attempts. This skill paired with a shared knowledge base allows the individuals and robots to build a reliable \emph{theory of mind} and collaborate effectively during various types of tasks \cite{matarese2022perception}.

Bera \etal \cite{beraModellingMultiChannelEmotions2019} proposed an emotion-aware navigation algorithm for social robots which combined emotions learned from facial expressions and walking trajectories using an onboard and an overhead camera respectively. The approach achieved accurate emotion detection and enabled socially conscious robot navigation in low-to-medium-density environments.

\begin{table*}[t]
	\centering
	\begin{threeparttable}
	\begin{tabularx}{\textwidth}{|m{0.05\textwidth}|X|X|X|X|X|X|X|X|X|X|X|X|X|X|X|}
		\toprule
		\multicolumn{2}{|c|}{} & \multicolumn{14}{c|}{\thead{Robots}} \\
		\midrule
		\multicolumn{2}{|c|}{} & 
		\rotatebox[origin=c]{90}{\thead{HRP-2}} & 
		\rotatebox[origin=c]{90}{\thead{HRP-4}} & 
		\rotatebox[origin=c]{90}{\thead{iCub}} & 
		\rotatebox[origin=c]{90}{\thead{NAO}} & 
		\rotatebox[origin=c]{90}{\thead{Sarcos}} & 
		\rotatebox[origin=c]{90}{\thead{Atlas}} & 
		\rotatebox[origin=c]{90}{\thead{Talos}} & 
		\rotatebox[origin=c]{90}{\thead{Valkyrie}} & 
		\rotatebox[origin=c]{90}{\thead{WALK-MAN}} & 
		\rotatebox[origin=c]{90}{\thead{Cassie}} &
		\rotatebox[origin=c]{90}{\thead{ARMAR-III}} &
		\rotatebox[origin=c]{90}{\thead{ARMAR-6}} &
		\rotatebox[origin=c]{90}{\thead{Romeo}} &
		\rotatebox[origin=c]{90}{\thead{Pepper}}
		\\
		\midrule
		\multirow{3}[30]{0.05\textwidth}{\rotatebox[origin=c]{90}{\thead{State \\ Estimation}}} 
		& \multicolumn{1}{l|}{\rotatebox[origin=c]{90}{\thead{Proprioceptive}}} 
		& \multicolumn{1}{l|}{\thead{\cite{Flayols17_simple}}} % hrp-2
		& \multicolumn{1}{l|}{\thead{\cite{Mori18_com_identification}}} % hrp-4
		& \multicolumn{1}{l|}{\thead{-}} % icub
		& \multicolumn{1}{l|}{\thead{\cite{Piperakis22_robust} \\ \cite{Piperakis18_nonlinear} \\ \cite{Piperakis16_zmp}}} % nao
		& \multicolumn{1}{l|}{\thead{\cite{Rotella14_state}\tnote{*}\\ \cite{Rotella15_momentum}\tnote{*}\\ \cite{Rotella16-inertial} }} % sarcos
		& \multicolumn{1}{l|}{\thead{\cite{Fallon14_drift}\\ \cite{Piperakis22_robust}\tnote{*} \\ 
				\cite{Xinjilefu14_decoupled} \\ \cite{Xinjilefu14_QP} \\ \cite{Manuelli16_externalforce} \\ \cite{Maravgakis23_probabilistic}\textsuperscript{*} \\ \cite{Xinjilefu15_com}}} % atlas
		& \multicolumn{1}{l|}{\thead{\cite{Piperakis22_robust} \\ \cite{Maravgakis23_probabilistic}}} % talos
		& \multicolumn{1}{l|}{\thead{\cite{Piperakis19_unsupervised}\tnote{*}}} % valkyrie
		& \multicolumn{1}{l|}{\thead{-}} % walk-man
		& \multicolumn{1}{l|}{\thead{\cite{Hartley20_contact}}} % cassie
		& \multicolumn{1}{l|}{\thead{-}} % armar-iii
		& \multicolumn{1}{l|}{\thead{-}} % armar-6
		& \multicolumn{1}{l|}{\thead{-}} % romeo
		& \multicolumn{1}{l|}{\thead{-}} % pepper
		\\
		\cmidrule{2-16}
		& \multicolumn{1}{l|}{\rotatebox[origin=c]{90}{\thead{Multi-sensor \\ filtering}}} 
		& \multicolumn{1}{l|}{\thead{-}} % hrp-2
		& \multicolumn{1}{l|}{\thead{-}} % hrp-4
		& \multicolumn{1}{l|}{\thead{-}} % icub
		& \multicolumn{1}{l|}{\thead{\cite{Piperakis19_outlier}}} % nao
		& \multicolumn{1}{l|}{\thead{-}} % sarcos
		& \multicolumn{1}{l|}{\thead{\cite{Fallon14_drift} \\ \cite{Camurri20Pronto}}} % atlas
		& \multicolumn{1}{l|}{\thead{-}} % talos
		& \multicolumn{1}{l|}{\thead{\cite{Camurri20Pronto}}} % valkyrie
		& \multicolumn{1}{l|}{\thead{\cite{Piperakis19_outlier}}} % walk-man
		& \multicolumn{1}{l|}{\thead{-}} % cassie
		& \multicolumn{1}{l|}{\thead{-}} % armar-iii
		& \multicolumn{1}{l|}{\thead{-}} % armar-6
		& \multicolumn{1}{l|}{\thead{-}} % romeo
		& \multicolumn{1}{l|}{\thead{-}} % pepper
		\\
		\cmidrule{2-16}
		& \multicolumn{1}{l|}{\rotatebox[origin=c]{90}{\thead{Multi-sensor \\ smoothing}}} 
		& \multicolumn{1}{l|}{\thead{\cite{Fourmy22}}} % hrp-2
		& \multicolumn{1}{l|}{\thead{-}} % hrp-4
		& \multicolumn{1}{l|}{\thead{-}} % icub
		& \multicolumn{1}{l|}{\thead{-}} % nao
		& \multicolumn{1}{l|}{\thead{-}} % sarcos
		& \multicolumn{1}{l|}{\thead{-}} % atlas
		& \multicolumn{1}{l|}{\thead{-}} % talos
		& \multicolumn{1}{l|}{\thead{-}} % valkyrie
		& \multicolumn{1}{l|}{\thead{-}} % walk-man
		& \multicolumn{1}{l|}{\thead{\cite{Hartley18_Hybrid} \\ \cite{Hartley18_legged}}} % cassie
		& \multicolumn{1}{l|}{\thead{-}} % armar-iii
		& \multicolumn{1}{l|}{\thead{-}} % armar-6
		& \multicolumn{1}{l|}{\thead{-}} % romeo
		& \multicolumn{1}{l|}{\thead{-}} % pepper
		\\
		\midrule
		\multirow{2}{0.05\textwidth}{\rotatebox[origin=c]{90}{\thead{Environment \\ Understanding}}} 
		& \multicolumn{1}{l|}{\rotatebox[origin=c]{90}{\thead{Localization, \\ and Navigation}}} 
		& \multicolumn{1}{l|}{\thead{-}} % hrp-2
		& \multicolumn{1}{l|}{\thead{
				\cite{zhang2019hrpslam} \\
				\cite{tanguy2019closed} \\
				\cite{zhang2020flowfusion}
			}} % hrp-4
		& \multicolumn{1}{l|}{\thead{-}} % icub
		& \multicolumn{1}{l|}{\thead{
				\cite{ovalle-magallanesTransferLearningHumanoid2021}
		 		\cite{ferroVisionBasedNavigationOmnidirectional2019} \\
 		 		\cite{regierClassifyingObstaclesExploiting2018}
	 		 	\cite{juangRobustVisualLinefollowing2020} \\
 		  		\cite{abiyevRobotPathfindingUsing2017}
 		  		\cite{osswaldEfficientCoverage3D2017} \\
	   			\cite{osswaldGPUAcceleratedNextBestViewCoverage2018},
	    		\cite{jinEnhancingBinocularDepth2018} \\
	    		\cite{monicaHumanoidRobotNext2019}
		     	\cite{wangFormationBuildingCollision2019} \\
		      	\cite{kaboliRobustTactileDescriptors2018a}}} % nao
		& \multicolumn{1}{l|}{\thead{-}} % sarcos
		& \multicolumn{1}{l|}{\thead{\cite{nobiliOverlapbasedICPTuning2017a}}} % atlas
		& \multicolumn{1}{l|}{\thead{-}} % talos
		& \multicolumn{1}{l|}{\thead{
				\cite{nobiliOverlapbasedICPTuning2017a} \\
				\cite{scona2017direct}
			}} % valkyrie
		& \multicolumn{1}{l|}{\thead{
				\cite{raghavanStudyLowDriftState2018} \\
				\cite{kanoulasFootstepPlanningRough2018}}} % walk-man
		& \multicolumn{1}{l|}{\thead{-}} % cassie
		& \multicolumn{1}{l|}{\thead{-}} % armar-iii
		& \multicolumn{1}{l|}{\thead{-}} % armar-6
		& \multicolumn{1}{l|}{\thead{\cite{magassoubaAuralServoSensorBased2018}}} % romeo
		& \multicolumn{1}{l|}{\thead{\cite{magassoubaAuralServoSensorBased2018}}} % pepper
		\\
		\cmidrule{2-16}
		& \multicolumn{1}{l|}{\rotatebox[origin=c]{90}{\thead{Grasping and \\ Manipulation}}} 
		& \multicolumn{1}{l|}{\thead{-}} % hrp-2
		& \multicolumn{1}{l|}{\thead{-}} % hrp-4
		& \multicolumn{1}{l|}{\thead{
				\cite{vezzaniGraspingApproachBased2017} \\
				\cite{vicenteMarkerlessVisualServoing2017} \\
				\cite{hoffmannRoboticHomunculusLearning2017} \\
				\cite{vezzaniMemoryUnscentedParticle2017}}} % icub
		& \multicolumn{1}{l|}{\thead{-}} % nao
		& \multicolumn{1}{l|}{\thead{-}} % sarcos
		& \multicolumn{1}{l|}{\thead{-}} % atlas
		& \multicolumn{1}{l|}{\thead{-}} % talos
		& \multicolumn{1}{l|}{\thead{-}} % valkyrie
		& \multicolumn{1}{l|}{\thead{\cite{nguyenObjectbasedAffordancesDetection2017}}} % walk-man
		& \multicolumn{1}{l|}{\thead{-}} % cassie
		& \multicolumn{1}{l|}{\thead{\cite{schmidtGraspingUnknownObjects2018}}} % armar-iii
		& \multicolumn{1}{l|}{\thead{
				\cite{vezzaniMemoryUnscentedParticle2017} \\ \cite{hundhausenFastReactiveGrasping2021}}} % armar-6
		& \multicolumn{1}{l|}{\thead{-}} % romeo
		& \multicolumn{1}{l|}{\thead{-}} % pepper
		\\
		\midrule
		\multirow{2}{0.05\textwidth}{\rotatebox[origin=c]{90}{\thead{Human-Robot \\ Interaction}}} 
		& \multicolumn{1}{l|}{\rotatebox[origin=c]{90}{\thead{Detection}}} 
		& \multicolumn{1}{l|}{\thead{\cite{yan2014survey}}} % hrp-2
		& \multicolumn{1}{l|}{\thead{-}} % hrp-4
		& \multicolumn{1}{l|}{\thead{\cite{davila-chaconEnhancedRobotSpeech2019} \\ \cite{yan2014survey} \\ \cite{matarese2022perception} \\ \cite{robinson2023robotic}}} % icub
		& \multicolumn{1}{l|}{\thead{\cite{schymuraExtendingLinearDynamical2018} \\ \cite{robinson2023robotic}}} % nao
		& \multicolumn{1}{l|}{\thead{\cite{robinson2023robotic}}} % sarcos
		& \multicolumn{1}{l|}{\thead{-}} % atlas
		& \multicolumn{1}{l|}{\thead{-}} % talos
		& \multicolumn{1}{l|}{\thead{-}} % valkyrie
		& \multicolumn{1}{l|}{\thead{-}} % walk-man
		& \multicolumn{1}{l|}{\thead{-}} % cassie
		& \multicolumn{1}{l|}{\thead{\cite{yan2014survey}}} % armar-iii
		& \multicolumn{1}{l|}{\thead{-}} % armar-6
		& \multicolumn{1}{l|}{\thead{-}} % romeo
		& \multicolumn{1}{l|}{\thead{\cite{moller2021survey} \\ \cite{robinson2023robotic}}} % pepper
		\\
		\cmidrule{2-16}
		& \multicolumn{1}{l|}{\rotatebox[origin=c]{90}{\thead{Inference}}} 
		& \multicolumn{1}{l|}{\thead{-}} % hrp-2
		& \multicolumn{1}{l|}{\thead{-}} % hrp-4
		& \multicolumn{1}{l|}{\thead{\cite{matarese2022perception}}} % icub
		& \multicolumn{1}{l|}{\thead{-}} % nao
		& \multicolumn{1}{l|}{\thead{-}} % sarcos
		& \multicolumn{1}{l|}{\thead{-}} % atlas
		& \multicolumn{1}{l|}{\thead{-}} % talos
		& \multicolumn{1}{l|}{\thead{-}} % valkyrie
		& \multicolumn{1}{l|}{\thead{-}} % walk-man
		& \multicolumn{1}{l|}{\thead{-}} % cassie
		& \multicolumn{1}{l|}{\thead{-}} % armar-iii
		& \multicolumn{1}{l|}{\thead{-}} % armar-6
		& \multicolumn{1}{l|}{\thead{-}} % romeo
		& \multicolumn{1}{l|}{\thead{\cite{beraModellingMultiChannelEmotions2019}}} % pepper
		\\
		\midrule
	\end{tabularx}
	\begin{tablenotes}\footnotesize
	\item[*] In simulation.
	\end{tablenotes}
	\end{threeparttable}
	\caption{A non-exhaustive, indicative list of popular humanoid robot models used by different publications.} 
	\label{tab:humanoid-references}
\end{table*}

%\section{Discussion}\label{sec12}
%
%Discussions should be brief and focused. In some disciplines use of Discussion or `Conclusion' is interchangeable. It is not mandatory to use both. Some journals prefer a section `Results and Discussion' followed by a section `Conclusion'. Please refer to Journal-level guidance for any specific requirements. 

\section{Conclusion}\label{sec13}

Substantial progress have been made in all three principal areas discussed in this survey. In \tabref{tab:humanoid-references} we compile a list of the most commonly cited humanoids in the literature corresponding to the aforementioned categorization. We conclude with a summary of the trends and possible areas of further research we observed in each of these areas. 

\paragraph{State Estimation} 

Tightly-coupled formulation of state estimation based on \map seems to be promising for future works as it offers several advantages, such as modularity and enabling seamless integration of new sensor types, and extending generic estimators with accommodating a wider range of perception sources in order to develop a whole-body estimation framework. By integrating high-rate control estimation and non-drifting localization based on \slam, this framework could provide real-time estimation for locomotion control purposes, and facilitate gait and contact planning.

Another important area of focus is the development of multi-contact detection and estimation methods for arbitrary unknown contact locations. By moving beyond rigid segment assumptions for humanoid structure and augmenting robots with additional sensors, such as strain gauges to directly measure segment deflections; the multi-contact detection and compensating for modeling errors can lead to more accurate state estimation and improved human-robot interactions.

%With the advent of more capable inference hardware, we see an increasing application of learning techniques in areas of localization, object identification and mapping where it is replacing handcrafted feature descriptors. However, visual classifiers such as CNNs struggle to identify unstructured \textquote{stuff} compared regularly shaped \textquote{objects}, leading to the use of memory intensive representations such as point clouds, and the need for improving classifier capabilities. 
%In the area of \slam which has many capable solutions in static environments, we see the research focusing on handling dynamic obstacles where multi-sensor fusion finds favor as the varying modalities increases robustness. Though scalability and real-time capability remains a challenge as wrangling multiple data streams over long sequences could easily overwhelm a humanoid's onboard computer. 
%Footstep planning exhibits trends towards rapid environment modeling from perception information for quick response to perturbation. However, consistent modeling of dynamic obstacles is still a largely open challenge. 
%Manipulation and long term global planning also leans towards the use of learning techniques to adapt to unforeseen constraints in the environment. But these methods rely heavily on learning representations or embeddings of high dimensional interactions between various perceived elements for reducing complexity. However, this is a challenging task and affords avenue for finding more capable methods which can express these relationships with greater efficiency, scope and accuracy.

\paragraph{Environment Understanding} 

With the availability of improved inference hardware, learning techniques are increasingly being applied in localization, object identification, and mapping, replacing handcrafted feature descriptors. However, visual classifiers like CNNs struggle with unstructured \textquote{stuff} compared to regularly shaped objects, necessitating memory-intensive representations such as point clouds and the need for enhanced classifier capabilities. In the field of \slam, which has robust solutions for static environments, research is focused on handling dynamic obstacles by favoring multi-sensor fusion for increased robustness. Scalability and real-time capability remain challenging due to the potential overload of a humanoid's onboard computer from wrangling multiple data streams over long sequences. Footstep planning shows a trend towards rapid environment modeling for quick responses, but consistent modeling of dynamic obstacles remains an open challenge. Manipulation and long-term global planning also rely on learning techniques to adapt to unforeseen constraints, requiring representations or embeddings of high-dimensional interactions between perceived elements for complexity reduction. However, finding more efficient, comprehensive, and accurate methods to express these relationships is an ongoing challenge.

\paragraph{Human Robot Interaction} 

Research in the field of  HRI has focused on understanding human intent and emotion through various elements such as body pose, motions, expressions, audio cues, and behavior. Though this may seem natural and trivial from a human's perspective, it is often a very challenging task to incorporate the same into robotic systems. Despite considerable progress in the above approaches, the ever-changing and unpredictable nature of human interaction necessitates additional steps that incorporate concepts like shared autonomy and shared perception. In this context, contextual information and memory play a crucial role in accurately perceiving the state and intentions of the humans with whom interaction is desired. Current research endeavors are actively focusing on these pivotal topics, striving to enhance the capabilities of humanoid robots in human-robot interactions while also considering trust, safety, explainability, and ethics during these interactions.

\backmatter

%\bmhead{Supplementary information}
%
%If your article has accompanying supplementary file/s please state so here. 
%
%Authors reporting data from electrophoretic gels and blots should supply the full unprocessed scans for key as part of their Supplementary information. This may be requested by the editorial team/s if it is missing.
%
%Please refer to Journal-level guidance for any specific requirements.

\bmhead{Acknowledgments}

This work has partially been funded by the Deutsche Forschungsgemeinschaft (DFG, German Research Foundation) under BE 4420/4-1 within the FOR 5351 – 459376902 – AID4Crops and under Germany’s Excellence Strategy, EXC-2070 – 390732324 – PhenoRob. 

\section*{Declarations}

\begin{itemize}
\item \textbf{Conflict of interest} The authors declare no competing interests.
\item \textbf{Human and Animal Rights and Informed Consent} This article does not contain any studies with human or animal subjects performed by any of the authors.
%\item Consent for publication
%\item Availability of data and materials
%\item Authors' contributions
\end{itemize}

\bibliography{sn-bibliography}% common bib file

\end{document}